\documentclass[10pt,twocolumn,letterpaper]{article}

\usepackage{cvpr}              

\newcommand{\ul}[1]{\underline{#1}}

\usepackage{booktabs}
\usepackage{multirow}
\usepackage{adjustbox}
\usepackage{float}

\definecolor{cvprblue}{rgb}{0.21,0.49,0.74}
\usepackage[pagebackref,breaklinks,colorlinks,allcolors=cvprblue]{hyperref}

\def\paperID{17810} 
\def\confName{CVPR}
\def\confYear{2025}

\title{\textsc{SemAlign3D}: Semantic Correspondence between RGB-Images through Aligning 3D Object-Class Representations}

\author{
Krispin Wandel \\
Shanghai Jiao Tong University \\
{\tt\small krispin.wandel@sjtu.edu.cn}
\and
Hesheng Wang\thanks{Corresponding author.} \\
Shanghai Jiao Tong University \\
{\tt\small wanghesheng@sjtu.edu.cn}
}

\begin{document}
\maketitle
\begin{abstract}

Semantic correspondence made tremendous progress through the recent advancements of large vision models (LVM). While these LVMs have been shown to reliably capture local semantics, the same can currently not be said for capturing global geometric relationships between semantic object regions. This problem leads to unreliable performance for semantic correspondence between images with extreme view variation. In this work, we aim to leverage monocular depth estimates to capture these geometric relationships for more robust and data-efficient semantic correspondence. First, we introduce a simple but effective method to build 3D object-class representations from monocular depth estimates and LVM features using a sparsely annotated image correspondence dataset. Second, we formulate an alignment energy that can be minimized using gradient descent to obtain an alignment between the 3D object-class representation and the object-class instance in the input RGB-image. Our method achieves state-of-the-art matching accuracy in multiple categories on the challenging SPair-71k dataset, increasing the PCK@0.1 score by more than 10 points on three categories and overall by 3.3 points from 85.6\% to 88.9\%. Additional resources and code are available at https://dub.sh/semalign3d.

\end{abstract}    
\section{Introduction}
\label{sec:intro}

\begin{figure}[t]
  \centering
   \includegraphics[width=1.0\linewidth]{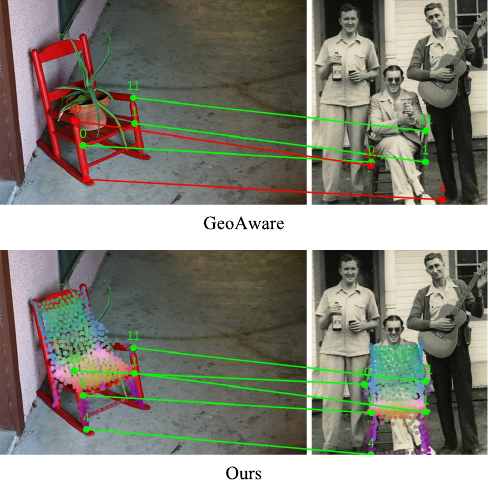}

   \caption{Our method aligns learned 3D object-class representations, displayed as the colored point cloud, with object instances to improve robustness of sparse semantic correspondence against extreme view variation and occlusion, which is still a major issue for existing methods like GeoAware \cite{24_left_right}.}
   \label{fig:hero}
\end{figure}

Semantic correspondence aims at finding an alignment between two images based on the meaning or function of image regions rather than their exact spatial or visual similarity. This ability is crucial for robotics to learn policies with category-level generalization capabilities \cite{18_dense_obj_nets, 19_cirr_visuo, 19_kpam, 21_kpam2, 22_nerf_dense}. Beyond robotics, this task has also several applications in image editing \cite{23_asic, 23_dragon, 23_neuCong, 23_sd+dinov2} and style transfer \cite{20_colorization, 23_tokenflow}. 

Large Vision Models (VIT) such as DinoV2 \cite{23_dinov2} have pushed this field dramatically through the introduction of deep features that are able to capture the underlying semantics of an object. However, as observed by \cite{24_left_right, 24_spherical_maps}, these features still fall short when dealing with symmetric objects or extreme view variation. Recent work has proposed solutions such as feature fine-tuning or the introduction of spatial bias (see \cref{sec:rel_work}) but the same issues still persist in many instances, as depicted in \cref{fig:hero}.

To derive a more robust method, we propose to learn category-level 3D object representations and then align these representations with the object instance shown in the image at inference time. Our hypothesis is that this process is significantly more data-efficient and robust than purely relying on features built at inference time. Towards this goal we make two contributions:
\begin{enumerate}
    \item We introduce a simple method to combine monocular depth estimates and deep features to build 3D object-class representations using a sparsely annotated dataset. (see \cref{chap:representation}).
    \item We define an alignment loss function that can be minimized using gradient descent to obtain the alignment between the 3D representation and the object instance shown in the image. (see \cref{chap:alignment}).
\end{enumerate}

Our method shows substantial performance gains over the previous state-of-the-art \cite{24_left_right} in multiple categories on the \textsc{SPair-71k} \cite{19_spair} dataset, increasing the PCK@0.1 score by  more than 10 points in three categories and the overall score by 3.3 points (from 85.6\% to 88.9\%) (see \cref{sec:results} for details).

\section{Related Work}
\label{sec:rel_work}



The field of semantic correspondence focuses on finding meaningful correspondences between parts of objects in different images, often within an unsupervised or self-supervised framework. Several approaches have been developed to address this, categorized broadly into learning mappings, optimizing correlation matrices, and fine-tuning feature representations.

\paragraph{Learning Mappings to Canonical Coordinates.} Unsupervised methods in this category, like ASIC \cite{23_asic} and NeuCongeal \cite{23_neuCong}, aim to learn mappings between deep features and canonical coordinates. ASIC minimizes contrastive and reconstruction losses, mapping features to canonical coordinates for both sparse and dense matching. NeuCongeal follows a similar approach but leverages deep feature-based reconstruction loss. Another notable approach, Viewpoint-Guided Spherical Maps \cite{24_spherical_maps}, maps features to canonical spherical coordinates, integrating a weak geometric prior, though it faces challenges with complex or non-rigid objects.

\paragraph{Optimizing the Correlation Matrix.} Techniques such as SCOT \cite{20_transport} and PMNC \cite{21_pmnc} optimize correlation matrices to improve matching precision. SCOT frames semantic correspondence as an optimal transport problem, adding geometric consistency with regularized Hough Matching (RHM). However, RHM can be inaccurate for symmetrical objects. PMNC refines correlation matrices via learned patch scoring functions. CATs++ \cite{22_cats++} combines multi-layer correlations using transformers, enhancing efficiency with convolutions.

\paragraph{Fine-Tuning Deep Features.} Recent works like SCorrSAN \cite{22_sparse, 16_dense_corr} employ sparse annotations and generate dense pseudo labels to improve keypoint matching accuracy. Other studies utilize features extracted from models like DinoV2 \cite{23_dinov2, 23_sd+dinov2} or stable diffusion (SD) \cite{23_sd, 24_sd4match}, with methods like DHF \cite{23_dhf} refining feature representations through cosine similarity loss. Fine-tuning DinoV2 and SD features across datasets with similar domains also demonstrates improved performance \cite{24_left_right}, \eg by fine-tuning on bird datasets to enhance airplane features.




 




\section{Method}
\label{sec:method}

\subsection{Building 3D Object-Class Representations}
\label{chap:representation}

\begin{figure*}
  \centering
  \includegraphics[width=0.9\linewidth]{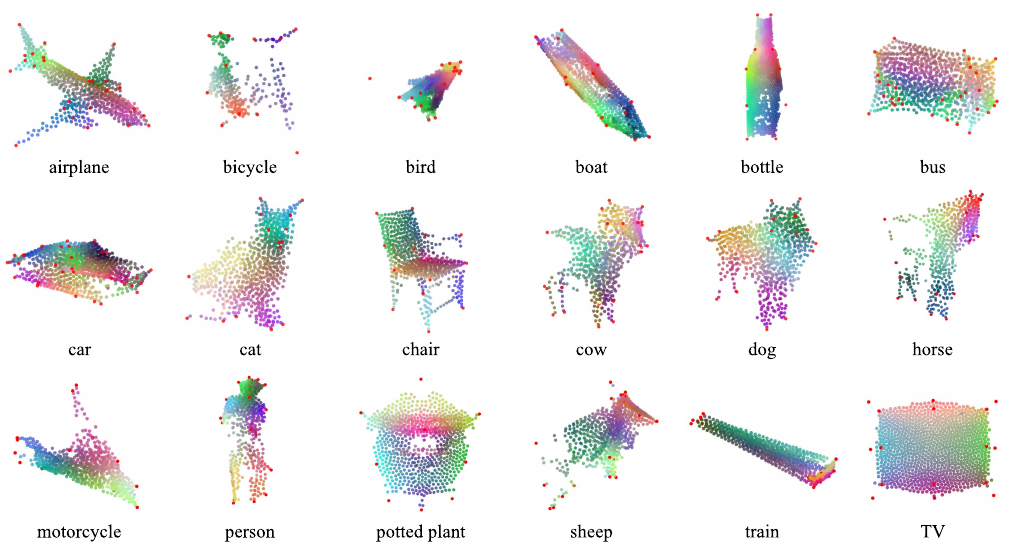}
  \caption{Learned 3D model representations for all SPair-71k \cite{19_spair} categories. Colors represent the principal components of the point cloud features.}
  \label{fig:dense_pc}
\end{figure*}

We aim to leverage the recent progress in monocular depth estimation \cite{24_depth_any, 24_depth-pro, 23_miragold} to build 3D object class representations from a small set of sparsely annotated RGB images. Specifically, we assume a set of $n$ images that display instances of an object class (\eg passenger airplane), and each image $s$ is sparsely annotated with $L$ labeled keypoints $k_{i=1\dots L, \text{im}}^s$ (\eg nose, left/right wing end). For notational simplicity, we assume all images have the same number of keypoints. Our goal is to obtain 3D representations that capture the semantic regions and their geometric relationships for each object category. Importantly, these representations do not embody a specific object (\eg Airbus 380), but a class of objects (\eg passenger airplane). In the following, we propose a simple but effective procedure to construct these 3D object class representations.

\paragraph{Keypoint World Coordinates.} First, we compute world coordinates for the image keypoints using monocular depth estimates from DepthAnythingV2 \cite{24_depth_any}. Since camera intrinsics are unknown, we approximate them up to scale by estimating focal length parameters. Given a focal length estimate, we can back-project 2D keypoints to 3D world coordinates and construct scale-invariant geometric features:
\begin{align}
  A_{ijkl}^s &= \angle (e_{ij}^s, e_{kl}^s) \\
  R_{ijkl}^s &= \frac{|e_{ij}^s|}{|e_{ij}^s| + |e_{kl}^s|},
  \label{eq:angle_ratio}
\end{align}
where $e_{ij}^s = k_{j, \text{world}}^{s} - k_{i, \text{world}}^{s}$ represents the edge between keypoints. We optimize focal length estimates by minimizing the variance of these features across $n$ train images:
\begin{equation}
    f_{1 \dots n}^* = \min_{f_{1 \dots n}} \sum_{ijkl}{ \mathop{\operatorname{Var}}_{s=1 \dots n}(R_{ijkl}^s) + \mathop{\operatorname{Var}}_{s=1 \dots n}(A_{ijkl}^s)}
\end{equation}

\begin{figure}
  \centering
   \includegraphics[width=1.0\linewidth]{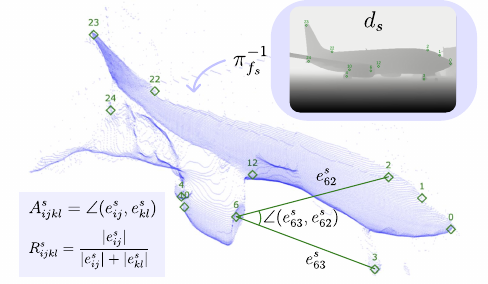}

   \caption{Construction of Geometric Features. Optimal focal lengths $f_s^*$ can be found by minimizing variance of scale-invariant geometric features $A_{ijkl}^{s=1 \dots n}$ and $R_{ijkl}^{s=1 \dots n}$. We use the state-of-the-art model DepthAnythingV2 \cite{24_depth_any} for monocular depth estimation. Although the produced depth map (top right) looks visually appealing, it is still far from perfect as evident when back-projected to world coordinates. Nevertheless, in this work we demonstrate that we can still use these depth maps to build coherent 3D object-class representations.}
   \label{fig:geom_fts}
\end{figure}

\begin{figure}
  \centering
   \includegraphics[width=3.25in]{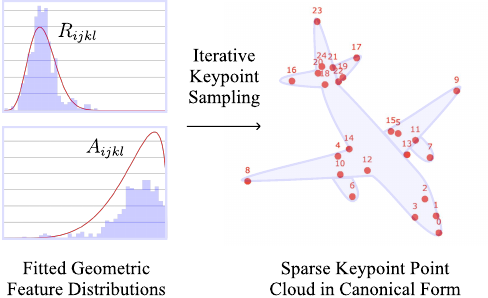}

   \caption{Construction of Sparse Keypoint Point Cloud in Canonical Form. Built by iteratively computing the next most likely keypoint location based on the fitted Beta-distributions over $A_{ijkl}^{s=1 \dots n}$ and $R_{ijkl}^{s=1 \dots n}$.}
   \label{fig:sparse_pc}
\end{figure}

\paragraph{Sparse 3D Representation.} Next, we construct a sparse \textit{canonical} point cloud $C_\text{sparse} = \{k_{i,\text{world}}\ |\ i=1\dots L\}$ by iteratively computing the most likely 3D location for each keypoint based on the Beta distributions $A_{ijkl}$ and $R_{ijkl}$ fitted to the geometric features on $n$ train images $A_{ijkl}^{s=1 \dots n}$ and $R_{ijkl}^{s=1 \dots n}$, respectively. For semantics, we extract image-patch features using the pre-trained model from GeoAware \cite{24_left_right}. Subsequently, we compute the semantic feature vector $q_i$ of $k_{i, \text{world}}$ as the average of $q_i^{s=1 \dots n}$, where $q_i^s$ denotes the image patch feature closest to $k_{i, \text{img}}^s$. In addition, we also compute the mean $a_{\mu_{i}}$ and the standard deviation $a_{\sigma_i}$ of the cosine similarities $a(q_i, q_i^{s=1 \dots n})$ between $q_i$ and $q_i^{s=1 \dots n}$ to obtain a keypoint presence certainty measure for $k_{i, \text{world}}$.

\paragraph{Dense 3D Representation.} Let $p_s = \pi_{f_s}^{-1}(d_s)$ denote the point cloud obtained by back-projecting the estimated depth image $d_s$. First, we align $p_s$ with the sparse canonical point cloud $C_\text{sparse}$ via barycentric parametrization with respect to the labeled keypoints $k_{i, \text{world}}^s$ and merge all aligned point clouds $p_{\text{aligned}}^{s=1 \dots n}$ into one large point cloud $p_\text{merged}$, while discarding all points that are outside of the convex hull spawned by the image keypoint world coordinates. Following this, we obtain the dense point cloud $C_\text{dense}$ through k-mean clustering $p_\text{merged}$, keeping only the clusters that are above a given density threshold. For semantics, let $\mathop{N}(c_i)$ of a point $c_i \in C_\text{dense}$ be the index set of the $m$ nearest neighbors to cluster $c_i$ in $p_\text{merged}$. Furthermore, let $q_j^s$ denote the image-patch feature that is closest to the image coordinate of a point $j$ in $N(c_i)$. Then, the semantic feature vector $q_{c_i}$ of $c_i \in C_\text{dense}$ is the average over the neighbor features $q_{\mathop{N}(c_i)} = \{q_j^s | j \in \mathbf{N}(c_i) \}$. As before, we also compute mean and variance of the cosine similarities between $q_{c_i}$ and $q_{\mathop{N}(c_i)}$ for presence certainty estimation.

\subsection{Aligning 3D Model Representations}
\label{chap:alignment}

\begin{figure*}[t]
  \centering
  \includegraphics[width=1.0\linewidth]{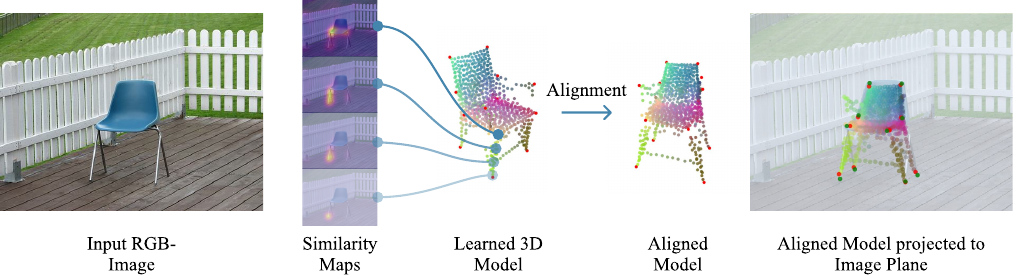}
  \caption{Overview of our method. After we constructed the object-class representation, we can minimize $\mathcal{L}_\text{align}$, which is based on similarity maps (here we  only show 4) between the input RGB-image and the representation, to obtain the alignment. The green dots in the final image on the right represent the ground-truth keypoints.}
  \label{fig:method_align}
\end{figure*}

We now focus on the core of our method: utilizing the dense 3D object-class representations for semantic correspondence. Specifically, given an RGB image with an object instance, our goal is to determine the optimal 3D keypoint coordinates ${k}_i^* \in {C}_\text{sparse}^*$ and focal length $f^*$ that yield the deformed dense point cloud ${C}_\text{dense}^*$ aligning precisely with the instance when projected to the image plane. Recall that ${C}_\text{dense}$ is parametrized by the sparse keypoint set ${C}_\text{sparse}$ through barycentric coordinates, allowing for flexible deformation. To achieve this alignment, we formulate a simple yet effective loss function, $\mathcal{L}_\text{align}({C}_\text{sparse}, f)$, minimized via gradient descent. This loss function consists of four key components: a  reconstruction loss that maximizes image likelihood, a geometric consistency loss to enforce structural plausibility, an optional background mask penalization term, and a regularization loss for faster convergence during optimization. Each of these loss terms will be explained in detail below.


\paragraph{Image Likelihood Maximization.} We define a likelihood function over image pixel coordinates that is maximal when the 3D class representation is well-aligned with the object instance. Let $q_\text{img}$ be the image embedding returned by the pre-trained GeoAware model \cite{24_left_right}. Then, we can define the semantic and spatial likelihood at $(x,y)$ for point $i$ in $C_\text{dense}$ as:
\begin{align}
p_{\text{sem}}^{xyi} &= g\left(a^\text{up}_{xy}(q_\text{img}, q_i);\ a_\mu^i, a_\sigma^i)\right) 
\label{e-sem} \\
p_\text{spatial}^{xyi}(C_\text{sparse}, f) &= g\left(|(x,y) - \pi_f(c_i)|;\ \mu=0, \sigma \right),
\label{e-spatial}
\end{align}
where $g$ is the max-normalized normal probability density function, $a^\text{up}$ is the cosine similarity up-scaled to image resolution, and $\sigma$ in \cref{e-spatial} is a hyper-parameter. As we explain later, during optimization, it is crucial to start with a large $\sigma$ and decrease it over time. The combined reconstruction likelihood is:
\begin{equation}
p_\text{reconstruct}^{xyi}(C_\text{sparse}, f) = p_\text{spatial}^{xyi}(f) \cdot p_{\text{sem}}^{xyi}
\label{e-p-reconstruct}
\end{equation}
Following this, the reconstruction loss is the negative total image likelihood defined as:
\begin{align}
P(\text{image} | C_\text{sparse}, f) &= \mathbf{E}_{x,y} \left [ \max_{c_i \in C_\text{dense}}{p_\text{reconstruct}^{xyi}} \right ] \\
\mathcal{L}_\text{reconstruct}^\text{dense} &= -P(\text{image} | C_\text{sparse}, f)
\label{e-p_image}
\end{align}
Since we maximize the expected value, in practice we do not optimize over all pixel coordinates but only over a few samples. Lastly, we also maximize the image likelihood with respect to the keypoints $k_i \in C_\text{sparse}$ to get a more precise position for the keypoints:
\begin{equation}
    \mathcal{L}_\text{reconstruct} = w_\text{dense} \cdot \mathcal{L}_\text{reconstruct}^\text{dense} + w_\text{sparse} \cdot \mathcal{L}_\text{reconstruct}^\text{sparse},
    \label{e-reconstruct}
\end{equation}
where $w_\text{dense}$ and $w_\text{sparse}$ are hyper-parameters. In practice, we start the optimization procedure with high $w_\text{dense}$ to avoid local minima and increase $w_\text{sparse}$ over time for improved precision.

\paragraph{Geometric Consistency.} To enforce geometric consistency during alignment, we maximize the likelihood of keypoint angles and edge ratios according to the fitted Beta distributions:
\begin{equation}
    \mathcal{L}_\text{geom} = -\mathbf{E}_{ijkl} \left[ A_{ijkl}(C_\text{sparse}) + B_{ijkl}(C_\text{sparse}) \right]
\end{equation}
It is worth noting that $C_\text{sparse}$ is initially already geometric consistent. This makes it much easier to optimize the loss function as we do not have to \textit{find} a configuration that is geometric consistent but only \textit{keep} the initial configuration geometric consistent.

\paragraph{Background Mask Penalization.} The reconstruction loss $\mathcal{L}_\text{reconstruct}$ tries to maximize the spatial-semantic likelihood for every pixel in the image. However, it does not penalize \textit{wrong positives}, that is, when the point cloud is projected to background image pixels that neither belong to the object segmentation mask $m_\text{seg}$ nor a region $m_\text{occ}$ that occludes the object. While we can obtain $m_\text{seg}$ by combining image attention with SegmentAnything \cite{23_seg_any}, obtaining $m_\text{occ}$ is very difficult. However, we observed that a background mask penalization loss $\mathcal{L}_\text{background}$ that is only based on $m_\text{seg}$ can still help to avoid local minima even when heavy occlusion is present. We define $\mathcal{L}_\text{background}$ as follows:
\begin{equation}
    \mathcal{L}_\text{background} = \frac{1}{|C_\text{dense}|} \sum_{c_i \in C_\text{dense}}\mathop{\mathbf{E}}_{xy \in m_\text{seg}} \left[ p_\text{spatial}^{xyi} \right]
\end{equation}

\paragraph{Regularization.} Since all features are scale invariant, we can constrain the motion of $C_\text{sparse}$ to boost convergence rate. Concretely, we add a soft constraint to the mean of the z-coordinates of $C_\text{sparse}$:
\begin{align}
    \mathcal{L}_\text{depth} &= (d_\text{target} - \mu_{\text{sparse}, z})^2,
\end{align}
where $d_\text{target}$ is a constant that can be set to an arbitrary value greater than zero.

\paragraph{Final Loss Function.} Putting it all together, the final alignment loss function is defined as:
\begin{equation}
\begin{split}
    \mathcal{L}_\text{align} &= w_\text{reconstruct} \cdot \mathcal{L}_\text{reconstruct} + w_\text{geom} \cdot \mathcal{L}_\text{geom} \\
    &\quad + w_\text{background} \cdot \mathcal{L}_\text{background} + w_\text{depth} \cdot \mathcal{L}_\text{depth}
\label{equ:loss_align}
\end{split}
\end{equation}
We will discuss the impact of each loss term and the choice of hyper-parameters $w_i$ in more detail in the next chapter.







\subsection{Semantic Correspondence}

After we found the optimal aligned point clouds $C_{\text{sparse},\{1,2\}}^*$ and focal lengths $f_{\{1,2\}}^*$ for images $s_1$ and $s_2$, we can perform semantic correspondence between $s_1$ and $s_2$ simply by traversing the alignment process in reverse order. That is, given a query image point $p_1$ in image $s_1$, we compute
\begin{equation}
    p_\text{reconstruct}^{p_1, i}(C_{\text{sparse},1}^*, f_1^*) = p_\text{spatial}^{p_1, i} \cdot p_\text{sem}^{p_1, i}
    \label{equ:rec_infer}
\end{equation}
similar to \cref{e-p-reconstruct} and choose the point index $i^*$ for which $p_\text{reconstruct}^{p_1, i}$ is maximal. Then, we can find its corresponding position in image $s_2$ by computing $p_\text{reconstruct}^{xy, i^*}(C_{\text{sparse},2}^*, f_2^*)$ for all image coordinates in $s_2$ and choosing the one with maximum likelihood. Another benefit of this approach is that we can adjust the variance of $p_\text{spatial}$ and $p_\text{sem}$ depending on how much we trust these terms.

\section{Experimental Evaluation}
\label{sec:results}

\begin{table*}[!ht]
\centering
\rowcolors{2}{gray!10}{white}
\begin{adjustbox}{width=\textwidth}
\begin{tabular}{llcccccccccccccccccccc}
\toprule
& \textbf{Method} & \textbf{Aero} & \textbf{Bike} & \textbf{Bird} & \textbf{Boat} & \textbf{Bottle} & \textbf{Bus} & \textbf{Car} & \textbf{Cat} & \textbf{Chair} & \textbf{Cow} & \textbf{Dog} & \textbf{Horse} & \textbf{Motor} & \textbf{Person} & \textbf{Plant} & \textbf{Sheep} & \textbf{Train} & \textbf{TV} & \textbf{All} \\ 
\midrule
\textbf{U}
\cellcolor{white} & ASIC \cite{23_asic} & 57.9 & 25.2 & 68.1 & 24.7 & 35.4 & 28.4 & 30.9 & 54.8 & 21.6 & 45.0 & 47.2 & 39.9 & 26.2 & 48.8 & 14.5 & 24.9 & 40.9 & 24.6 & 36.9 \\ 
\cellcolor{white} & DINOv2+NN \cite{23_dinov2, 23_sd+dinov2} & 72.7 & 62.0 & 85.2 & 41.3 & 40.4 & 52.3 & 51.5 & 71.1 & 36.2 & 67.1 & 64.6 & 67.6 & 61.0 & 68.2 & 30.7 & 62.0 & 54.3 & 24.2 & 55.6 \\ 
\cellcolor{white} & DIFT \cite{23_sd} & 63.5 & 54.5 & 80.8 & 34.5 & 46.2 & 52.7 & 48.3 & 77.7 & 39.0 & 76.0 & 54.9 & 61.3 & 53.3 & 46.0 & 57.8 & 57.1 & 71.1 & 63.4 & 57.7 \\ 
\cellcolor{white} & SD+DINOv2 \cite{23_sd+dinov2} & 73.0 & 64.1 & 86.4 & 40.7 & 52.9 & 55.0 & 53.8 & 78.6 & 45.5 & 77.3 & 64.7 & 69.7 & 63.3 & 69.2 & 58.4 & 67.6 & 66.2 & 53.5 & 64.0 \\ 
\cellcolor{white} & SphericalMaps \cite{24_spherical_maps} & 74.8 & 64.5 & 87.1 & 45.6 & 52.7 & 77.8 & 71.4 & 82.4 & 47.7 & 82.0 & 67.3 & 73.9 & 67.6 & 60.0 & 49.9 & 69.8 & 78.5 & 59.1 & 67.3 \\
\midrule
\textbf{S}
\cellcolor{white} & SCOT \cite{20_transport} & 34.9 & 20.7 & 63.8 & 21.1 & 43.5 & 27.3 & 21.3 & 63.1 & 20.0 & 42.9 & 42.5 & 31.1 & 29.8 & 35.0 & 27.7 & 24.4 & 48.4 & 40.8 & 35.6 \\ 
\cellcolor{white} & PMNC \cite{21_pmnc} & 54.1 & 35.9 & 74.9 & 36.5 & 42.1 & 48.8 & 40.0 & 72.6 & 21.1 & 67.6 & 58.1 & 50.5 & 40.1 & 54.1 & 43.3 & 35.7 & 74.5 & 59.9 & 50.4 \\ 
\cellcolor{white} & SCorrSAN \cite{22_sparse} & 57.1 & 40.3 & 78.3 & 38.1 & 51.8 & 57.8 & 47.1 & 67.9 & 25.2 & 71.3 & 63.9 & 49.3 & 45.3 & 49.8 & 48.8 & 40.3 & 77.7 & 69.7 & 55.3 \\ 
\cellcolor{white} & CATs++ \cite{22_cats++} & 60.6 & 46.9 & 82.5 & 41.6 & 56.8 & 64.9 & 50.4 & 72.8 & 29.2 & 75.8 & 65.4 & 62.5 & 50.9 & 56.1 & 54.8 & 48.2 & 80.9 & 74.9 & 59.8 \\ 
\cellcolor{white} & DHF \cite{23_dhf} & 74.0 & 61.0 & 87.2 & 40.7 & 47.8 & 70.0 & 74.4 & 80.9 & 38.5 & 76.1 & 60.9 & 66.8 & 66.6 & 70.3 & 58.0 & 54.3 & 87.4 & 60.3 & 64.9 \\ 
\cellcolor{white} & SD+DINOv2 (S) \cite{23_sd+dinov2} & 81.2 & 66.9 & 91.6 & 61.4 & 57.4 & 85.3 & 83.1 & 90.8 & 54.5 & 88.5 & 75.1 & 80.2 & 71.9 & 77.9 & 60.7 & 68.9 & 92.4 & 65.8 & 74.6 \\
\cellcolor{white} & GeoAware (AP-10k P.T.) \cite{24_left_right} & \ul{92.0} & \ul{76.1} & \textbf{97.2} & \textbf{70.4} & \ul{70.5} & \ul{91.4} & \ul{89.7} & \textbf{92.7} & \ul{73.4} & \ul{95.0} & \ul{90.5} & \textbf{87.7} & \ul{81.8} & \ul{91.6} & \ul{82.3} & \ul{83.4} & \textbf{96.5} & \ul{85.3} & \ul{85.6} \\ 
\cmidrule(lr){2-21}
\rowcolor{white}
\cellcolor{white} & \textbf{Ours} & \textbf{95.6} & \textbf{80.4} & \ul{94.0} & \ul{66.5} & \cellcolor{yellow} \textbf{82.2} & \textbf{92.6} & \textbf{92.5} & \ul{91.4} & \cellcolor{yellow} \textbf{88.3} & \textbf{96.1} & \textbf{91.3} & \ul{87.4} & \textbf{84.2} & \textbf{92.3} & \textbf{87.7} & \textbf{85.3} & \ul{95.9} & \cellcolor{yellow} \textbf{96.1} & \textbf{88.9} \\ 
\bottomrule
\end{tabular}
\end{adjustbox}
\caption{Comparison of PCK@0.1 accuracy for all categories in the \textsc{SPair-71k} dataset. \textbf{U}=unsupervised, \textbf{S}=supervised.}
\label{t-spair-cat}
\end{table*}

\subsection{Experimental Setup}

\paragraph{Datasets.} Following previous work in semantic correspondence, we evaluate our method on the \textsc{SPair-71k} dataset \cite{19_spair}. \textsc{SPair-71k} contains around 71,000 image pairs, with each pair belonging to the same object category, such as animals, vehicles, or household items. Each pair includes detailed annotations like keypoints and bounding boxes. The dataset features significant variations in viewpoint, scale, occlusion, and variation among objects within each category, making it one of the most challenging datasets for semantic correspondence. For completeness, we also evaluate our method on a small subset of \textsc{AP-10k} \cite{21_ap10k} and \textsc{PF-Pascal} \cite{16_pfpascal, 17_pfpascal}.

\paragraph{Evaluation Metrics.} For a fair comparison with previous work we follow common practice and use the Percentage of Correct Keypoints (PCK) \cite{12_pck} as primary evaluation metric for sparse semantic correspondence. PCK at level $\alpha$ is defined as the percentage of predicted points that are no further away than $\alpha \cdot \max(w_\text{bbox}, h_\text{bbox})$ from the ground truth, where $w_\text{bbox}$ and $h_\text{bbox}$ are the height and width of the corresponding object-instance bounding boxes. While PCK is a popular metric for semantic correspondence, it primarily measures accuracy, not precision. Therefore, we also report results for PCK\textsuperscript{\dag} \cite{22_pck+} and KAP \cite{24_spherical_maps}.

\paragraph{Optimization.} We start the optimization procedure with a random initial guess for $C_\text{sparse}$ and the focal length $f$. To evaluate $\mathcal{L}_\text{reconstruct}$ in \cref{e-p_image}, we sample a few image points based on the attention between the image and $Q_\text{dense}$. Although we observed that our method is good at finding the global minimum, there are initial configurations where the method can end up in a bad local minimum. To address this issue, we simply start with multiple initial guesses optimized in parallel. In our experiments, we start with $f \in [10.0, 5.0, 2.5, 1.25]$ and four $C_\text{sparse}$ samples per focal length. After optimization, instead of simply choosing the solution with the smallest loss, we add two post-processing steps: First, for object-class representations that are almost flat, such as TV or bottle, we apply a normal check and discard solutions that face away from the viewer. Second, we use the depth estimate of the image and penalize solutions when the z-order of the corresponding $C^*_\text{dense}$ is locally different from the depth estimate. 


\paragraph{Hyper-Parameters.} For the specific hyper-parameters per category please refer to the supplementary material. However, there are a few general important considerations for choosing the hyper-parameters:

First, it drastically improves performance when starting the optimization with a large $\sigma$ in $p_\text{spatial}$ from \cref{e-spatial} and gradually decreasing its value over time. Intuitively, this makes sense because our initial guess for $C_\text{sparse}$ might be quite off which should be reflected in the spatial likelihood estimate. Another side benefit of this scheduling policy is that it is very robust against image pixel outliers with wrong semantics because, as we decrease $\sigma$, the gradients with respect to these pixels will converge to zero. 

Another set of hyper-parameters includes the loss term weights in \cref{equ:loss_align}. Generally, we found that our method is not very sensitive with respect to these weights. However, one should choose $w_\text{geom}$ within reasonable bounds in order to make $C_\text{sparse}$ not too stiff. Furthermore, one should not start the optimization with a high value for $w_\text{depth}$, because this might push $C_\text{sparse}$ outside the image depending on the level of object-instance occlusion.

\subsection{Results and Analysis}

\begin{table}
\centering
\rowcolors{2}{gray!10}{white}
\begin{adjustbox}{width=\linewidth}
\begin{tabular}{llccc}
\toprule
& \textbf{Method} & \textbf{PCK@0.01} & \textbf{PCK@0.05} & \textbf{PCK@0.1} \\ 
\midrule
\textbf{U}
\cellcolor{white} & DINOv2+NN \cite{23_dinov2, 23_sd+dinov2} & 6.3 & 38.4 & 53.9 \\
\cellcolor{white} & DIFT \cite{23_sd} & 7.2 & 39.7 & 52.9 \\
\cellcolor{white} & SD+DINOv2 \cite{23_sd+dinov2} & 7.9 & 44.7 & 59.9 \\
\cellcolor{white} & SphericalMaps \cite{24_spherical_maps} & - & - & 69.8 \\
\midrule
\textbf{S}
\cellcolor{white} & SCorrSAN \cite{22_sparse} & 3.6 & 36.3 & 55.3 \\
\cellcolor{white} & CATs++ \cite{22_cats++} & 4.3 & 40.7 & 59.8 \\
\cellcolor{white} & DHF \cite{23_dhf} & 8.7 & 50.2 & 64.9 \\
\cellcolor{white} & SD+DINOv2 (S) \cite{23_sd+dinov2} \cite{23_sd+dinov2} & 9.6 & 57.7 & 74.6 \\
\cellcolor{white} & GeoAware (AP-10k P.T.) \cite{24_left_right} & \textbf{22.0} & \ul{75.3} & \ul{85.6} \\ 
\cmidrule(lr){2-5}
\cellcolor{white} & \textbf{Ours} & \ul{15.8} & \textbf{77.5} & \textbf{88.9}  \\  
\bottomrule
\end{tabular}
\end{adjustbox}
\caption{PCK scores for different levels averaged over all categories in the \textsc{SPair-71k} dataset. \textbf{U}=unsupervised, \textbf{S}=supervised.}
\label{t-spair-pck}
\end{table}


\begin{table}
\centering
\begin{adjustbox}{width=\linewidth}
\renewcommand{\arraystretch}{1.2}
\begin{tabular}{llccccccccc}
\toprule
& & \multicolumn{3}{c}{\textbf{PCK}} & \multicolumn{3}{c}{\textbf{PCK\textsuperscript{\dag}}} & \multicolumn{3}{c}{\textbf{KAP}} \\
\textbf{Dataset} & \textbf{Method} & $\alpha_1$ & $\alpha_2$ & $\alpha_3$ & $\alpha_1$ & $\alpha_2$ & $\alpha_3$ & $\alpha_1$ & $\alpha_2$ & $\alpha_3$ \\
\midrule
SPair-71k & GeoAware & 85.6 & 75.3 & \textbf{22.0} & \textbf{80.7} & \textbf{73.2} & \textbf{22.7} & 75.3 & 66.3 & \textbf{48.9} \\
& \textbf{Ours} & \textbf{88.9} & \textbf{77.5} & 15.8 & 80.5 & 72.5 & 15.4 & \textbf{76.9} & \textbf{67.3} & 46.6 \\
\rowcolor{gray!10} PF-Pascal & GeoAware & \textbf{89.6} & \textbf{84.8} & \textbf{72.0} & \textbf{83.2} & \textbf{79.7} & \textbf{69.4} & \textbf{86.5} & \textbf{81.7} & \textbf{70.1} \\
\rowcolor{gray!10} & \textbf{Ours} & 86.6 & 78.5 & 61.4 & 74.1 & 69.9 & 57.1 & 83.2 & 73.9 & 57.4 \\
AP-10k & GeoAware & \textbf{92.8} & \textbf{83.9} & \textbf{23.6} & \textbf{87.7} & \textbf{81.2} & \textbf{23.4} & \textbf{79.6} & \textbf{66.7} & \textbf{48.0} \\
& \textbf{Ours} & 89.1 & 68.1 & 7.3 & 82.3 & 64.9 & 7.2 & 77.4 & 58.8 & 40.4 \\
\bottomrule
\end{tabular}
\end{adjustbox}
\caption{Comparison of PCK, PCK\textsuperscript{\dag}, and KAP across datasets. We evaluated \textsc{AP-10k} on a small subset \{antelope, giraffe, elephant, hippo\} and used features fine-tuned on \textsc{SPair-71k} for PF-Pascal. $\alpha_{1,2,3}$ are [0.15, 0.1, 0.05] on \textsc{PF-Pascal} and [0.1, 0.05, 0.01] otherwise.}
\label{tab:more_eval}
\end{table}

\paragraph{Quantitative Analysis.} \Cref{t-spair-cat} presents PCK@0.1 results for each \textsc{SPair-71k} category, showing that our approach outperforms the state of the art in 13 of 18 categories, with comparable results in the remaining five. We achieve substantial improvements, including over 10 percentage points in Bottle, Chair, and TV categories, along with strong performance in other rigid object categories such as Airplane, Bike, and Plant. Our method demonstrates that even with simplistic representations, one can still achieve meaningful improvements. However, further enhancement of object-class representation learning is needed to improve performance on animals, where our method currently mainly benefits from feature averaging and certainty estimation. Overall, we see a 3.3\% increase (from 85.6\% to 88.9\%), closing the gap to 100\% by 22.9\%.

The overall PCK scores at different levels (0.01, 0.05, and 0.1) are displayed in \cref{t-spair-pck}. Our method shows a notable improvement of 2.2\% (from 75.3\% to 77.5\%) over GeoAware \cite{24_left_right} at the 0.05 level, though the gains are less pronounced than at the 0.1 level, with a slight drop in performance at the 0.01 level. This reduction may stem from the spatial bias in our approach, which could impact precision. We believe that refining model representations will help address this limitation and improve performance across all levels.

\Cref{tab:more_eval} includes additional metrics and results on \textsc{SPair-71k}, \textsc{AP-10k} \cite{21_ap10k}, and \textsc{PF-Pascal} \cite{16_pfpascal, 17_pfpascal} datasets, following GeoAware's evaluation protocol. Notably, the \textsc{AP-10k} test set does not include multi-instance cases. Our method does not perform as well here for several reasons. First, the spatial bias introduced through our method may negatively impact precision, as seen in results for PCK\textsuperscript{\dag} and KAP. Second, as we discuss later, our representations are not yet effective in capturing exact joint limits and symmetries, leading to weaker performance on non-rigid objects, explaining the results on \textsc{AP-10k}. However, we anticipate that joint-angle aware representations will improve performance, particularly in challenging cases like crossed legs in animals. Lastly, \textsc{PF-Pascal} favors traditional architectures due to pose similarities in source and target image, and test set images overlap with the training set. Our method excels in more realistic setups with high view variation.

\begin{figure*}[t]
  \centering
  \includegraphics[width=1.0\linewidth]{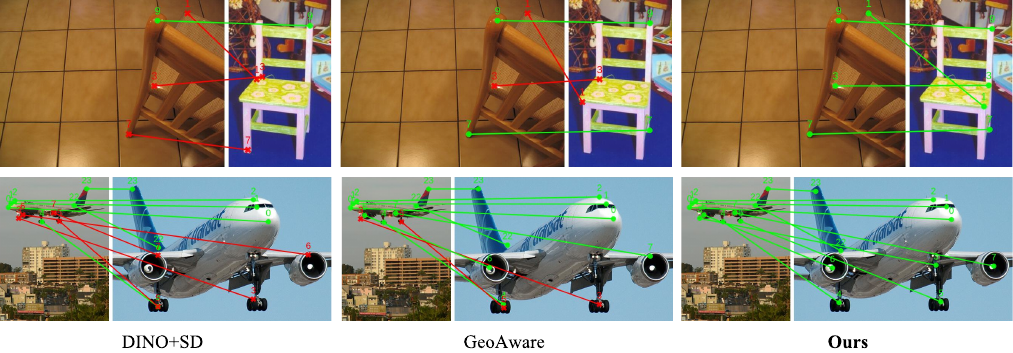}
  \caption{Qualitative comparison for sparse correspondence between DINO+SD \cite{23_sd+dinov2}, GeoAware (AP-10K P.T.) \cite{24_left_right}, and our method. Green lines (o-o) denote correct matches and red lines depict wrong matches (x-x).}
  \label{fig:sparse_matches}
\end{figure*}

\paragraph{Qualitative Analysis.} \Cref{fig:sparse_matches} illustrates sparse matches predicted by DINO+SD \cite{23_sd+dinov2}, GeoAware \cite{24_left_right}, and our method for two image pairs $(s_\text{src}^1, s_\text{tgt}^1)$ at the top and $(s_\text{src}^2, s_\text{tgt}^2)$ at the bottom, each presenting different challenges. In the first image pair, the chair is rotated by almost 180 degrees between $s_\text{src}^1$ and $s_\text{tgt}^1$, and in $s_\text{tgt}^2$ of the second image pair, the airplane's front wheel is aligned with the left wheel. GeoAware \cite{24_left_right} demonstrates that fine-tuning the backbone features of DINO+SD \cite{23_sd+dinov2}, which solely relies on combining unrefined DinoV2 \cite{23_dinov2} and Stable Diffusion features \cite{23_sd}, can lead to significant better results. However, it still struggles with extreme view variations, likely requiring substantial more data to learn robust features. In contrast, our method seems data-efficient, relying only on locally semantic information, such as ``wing-end", rather than ``\textit{left} wing end". Overall, qualitatively, our method shows a clear advantage in handling these challenging view changes.

\begin{figure}
  \centering
   \includegraphics[width=1.0\linewidth]{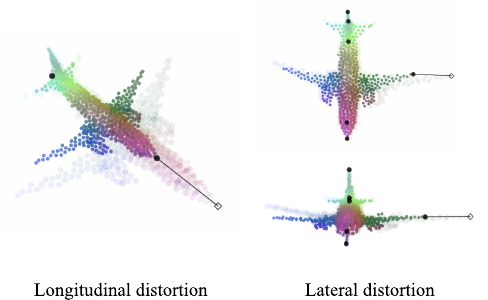}

   \caption{Behavior analysis of learned 3D object-class representation under external distortion. Black dots represent fixed points and the diamonds represent the target positions of fixed points that are being distorted. }
   \label{fig:distort}
\end{figure}

\paragraph{Representation Analysis. \label{par:representation}} As depicted in \cref{fig:distort}, we qualitatively examine the behavior of our 3D object-class representations by distorting a keypoint of the airplane model while keeping other points fixed and observe the resulting deformation. When the airplane is stretched longitudinally, it scales as expected. However, when the tip of the left wing is moved laterally outwards, we expect the right wing to move in the opposite direction to maintain symmetry. Instead, the right wing stays in place. Symmetry would lead to distributions over ratios with very little variance, but due to the noisy nature of the depth estimates, the variance is naturally larger. Therefore, there may not be enough force to push the end of the right wing outward. This suggests that there is still much room for improvement in learning object-class representations. We believe it should be possible to obtain high-quality object-class representations even with very little data available, similar to techniques like NERF \cite{20_nerf} for learning scene representations.


\begin{figure}
  \centering
   \includegraphics[width=1.0\linewidth]{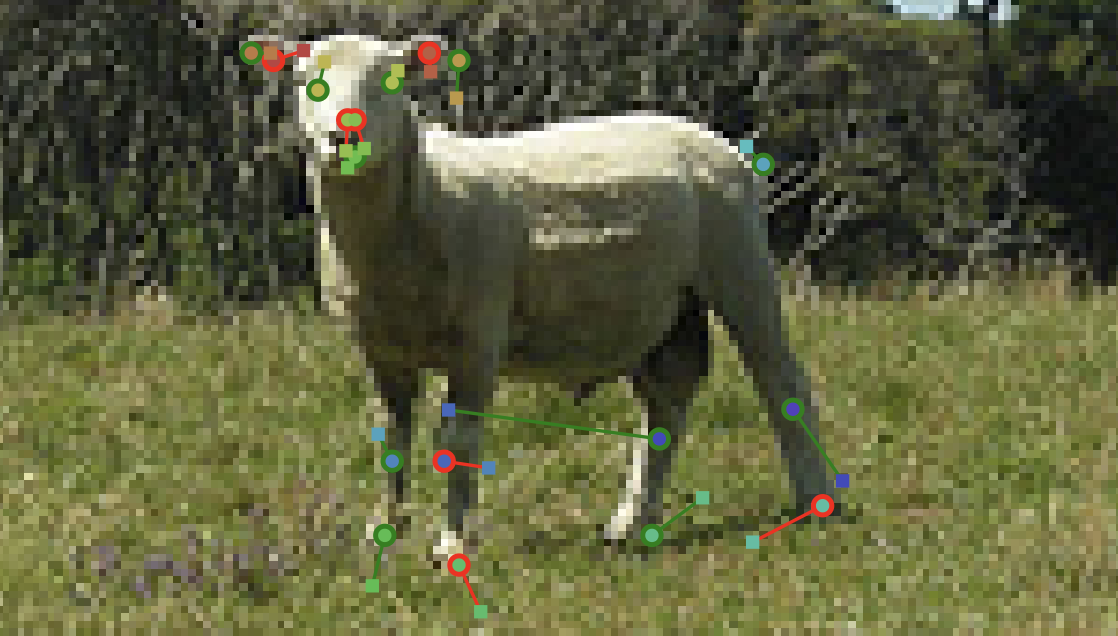}

   \caption{Failure Example for Sheep. Round markers represent ground truth keypoints. Square markers represent the keypoints in $C_\text{sparse}^*$ projected onto the image plane. Keypoints are sometimes mislabeled (red) when the semantic information is too inaccurate for the spatial likelihood to correct. Similarly, the spatial likelihood can also be inaccurate due to improper alignment.}
   \label{fig:failure}
\end{figure}

\paragraph{Failure Cases.} \Cref{fig:failure} depicts a failure case for the sheep category. The main cause of failure is misalignment between the object instance and the 3D representation. We expect that this problem can be solved by improving 3D object-class representations. Currently, if the alignment error is generally large for a specific category, we choose a large $\sigma$ for $p_\text{spatial}$ in \cref{equ:rec_infer} to mitigate this issue.

\paragraph{Ablation Studies}

\begin{table}
\centering
\rowcolors{2}{gray!10}{white}
\begin{adjustbox}{width=\linewidth}
\begin{tabular}{lccc}
\toprule
\textbf{Ablation} & \textbf{Bottle} & \textbf{Chair} & \textbf{TV} \\ 
\midrule
DinoV2+SD backbone features & 60.2 & 68.7 & 64.3 \\
Without $l_\text{reconstruct}^\text{sparse}$ & 79.1 & 84.2 & 92.1 \\
Without $l_\text{reconstruct}^\text{dense}$ & 64.0 & 63.0 & 85.9 \\
Without $l_\text{background}$ & 80.5 & 83.5 & 94.8 \\
Without $l_\text{depth}$ & 81.4 & 85.1 & 94.3 \\
\textbf{Original} & 82.2 & 88.3 & 96.1 \\
\bottomrule
\end{tabular}
\end{adjustbox}
\caption{PCK@0.1 scores of bottle, chair, and TV for various ablations.}
\label{tab:ablation}
\end{table}

In our ablation studies, summarized in \cref{tab:ablation}, we investigate the influence of the backbone features and the loss terms. We present the ablation results for the bottle, chair, and TV category on which the ablation effects are most pronounced, as we obtained the most performance gains in these categories. Evidently, switching from the fine-tuned backbone features \cite{24_left_right} to the unrefined DINOv2+SD \cite{23_sd+dinov2} features significantly decreases performance. Notably, however, we still perform better than any other unsupervised method. It can also be seen that both $\mathcal{L}_\text{reconstruct}^\text{sparse}$ and $\mathcal{L}_\text{reconstruct}^\text{dense}$ are crucial for optimal performance. Generally, $\mathcal{L}_\text{reconstruct}^\text{dense}$ seems more important than $\mathcal{L}_\text{reconstruct}^\text{sparse}$, underlying the significance of dense representations. Unsurprisingly, $\mathcal{L}_\text{background}$ and $\mathcal{L}_\text{depth}$ are not as critical as $\mathcal{L}_\text{reconstruct}$ but can still be helpful to get the best performance. Lastly, we did not hold out $\mathcal{L}_\text{geom}$ because this loss term is fundamental to our approach.







\subsection{Discussion}

\paragraph{Runtime.} Since we have to solve an optimization problem each time we process an image, our method is relatively slow compared to previous work. We found that it can take up to 10-30 seconds on a NVIDIA-3090 GPU to process a single image depending on the number of initial guesses and the number of points in $C_\text{sparse}$ and $C_\text{dense}$. That being said, there is likely a lot of potential for runtime optimization. However, in this work we did not focus on runtime because we are mainly interested in using our method for high-level planning for robotics, in which case we believe it is not prohibitive when the robot \textit{thinks} for a few more seconds before executing a command. Lastly, we think it is exciting to explore trading off inference time for better results with less data.

\paragraph{Limitations and Future Work.} Our method is mainly limited by the expressiveness of the 3D object-class representations. However, our hypothesis is that learning high-quality 3D object-class representations and then aligning them at inference time is significantly more data efficient and robust than purely relying on features built at inference time. This work showed that substantial performance gains can be achieved even with very simplistic 3D object-class representations, supporting our hypothesis.

Another limitation of our method is that it currently does not support end-to-end learning as representation learning  and alignment are two separated processes at the moment. However, it would be interesting to explore making the pipeline fully differentiable and adapting it to the unsupervised setting.

\section{Conclusion}
\label{sec:conclusion}

We introduce \textsc{SemAlign3D}, a novel approach for achieving robust semantic correspondence by aligning 3D object-class representations with RGB images. Leveraging monocular depth estimates and features from large vision models, we construct 3D representations that capture the semantics and geometric relations within an object class. Our method, demonstrated on the \textsc{SPair-71k} dataset, outperforms existing techniques by significantly enhancing matching accuracy, particularly on rigid object categories with extreme viewpoint variations. These results underscore the potential of our approach for data-efficient and robust semantic correspondence and we hope inspires future research on learning 3D object-class representations and alignment.

{
    \small
    \bibliographystyle{ieeenat_fullname}
    \bibliography{main}
}

\section*{Acknowledgement} This work was supported in part by the Natural Science Foundation of China under Grants 62225309, U24A20278, 62361166632, and U21A20480.

\clearpage
\setcounter{page}{1}
\maketitlesupplementary
\appendix
\renewcommand{\thesection}{\Alph{section}} 

\section{Additional Experimental Results}
\label{sec:results_suppl}

\begin{figure*}[t]
  \centering
  \includegraphics[width=0.85\linewidth]{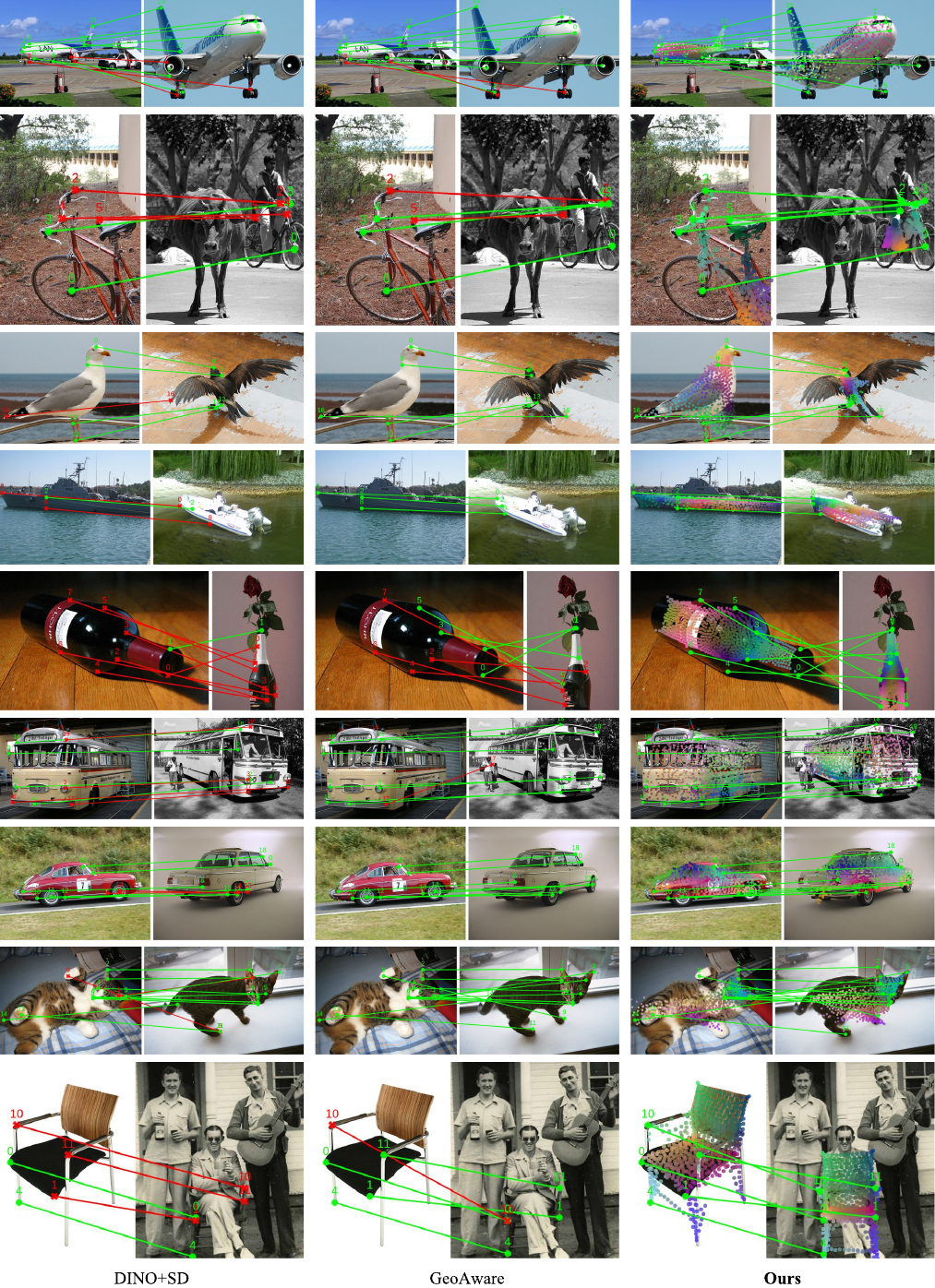}
  \caption{Qualitative comparison on SPair-71k \cite{19_spair} categories for sparse correspondence between DINO+SD \cite{23_sd+dinov2}, GeoAware (AP-10K P.T.) \cite{24_left_right}, and our method. Green lines (o-o) denote correct matches and red lines depict wrong matches (x-x).}
  \label{fig:sparse_matches_sup1}
\end{figure*}

\begin{figure*}[t]
  \centering
  \includegraphics[width=0.9\linewidth]{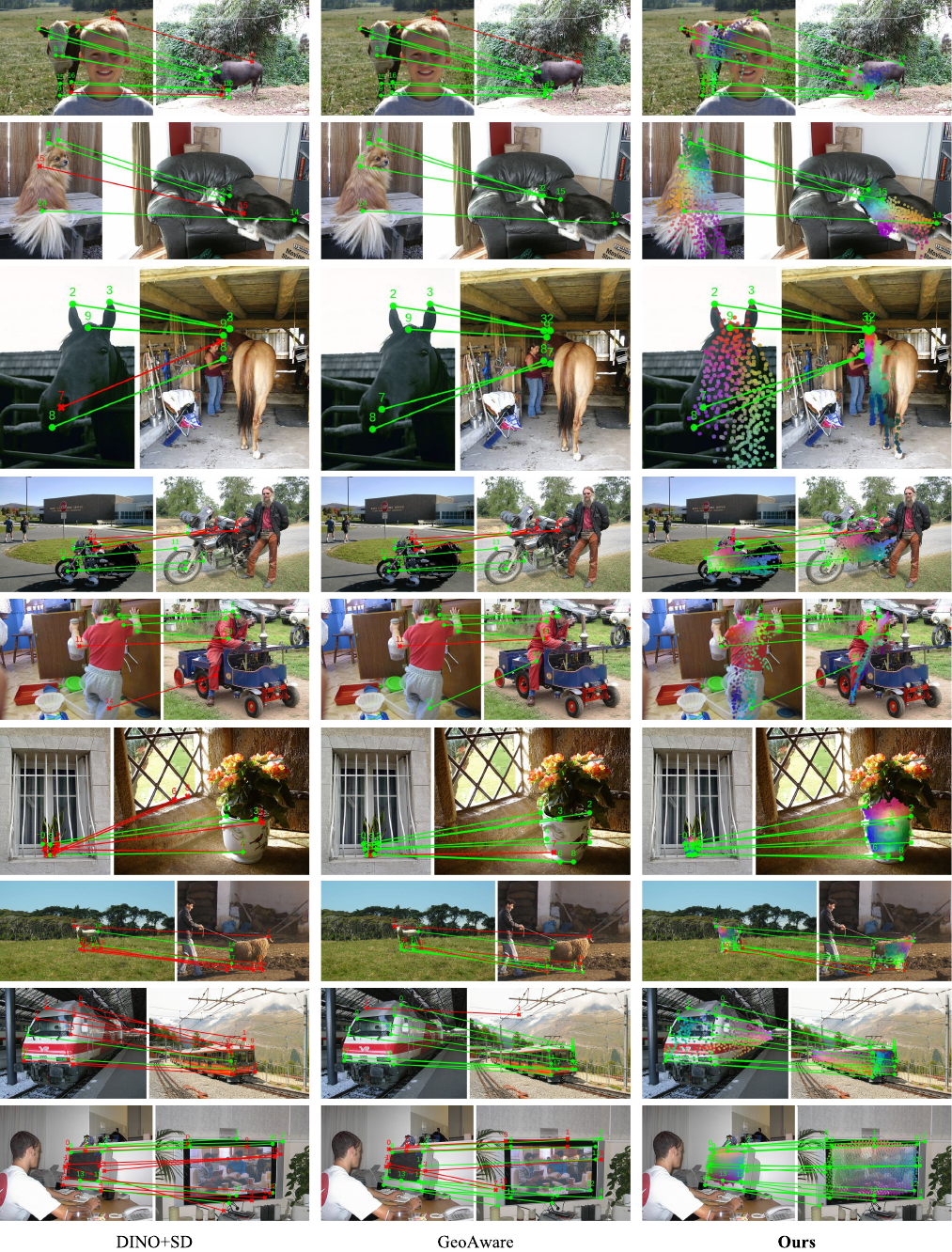}
  \caption{Qualitative comparison on SPair-71k \cite{19_spair} categories for sparse correspondence between DINO+SD \cite{23_sd+dinov2}, GeoAware (AP-10K P.T.) \cite{24_left_right}, and our method. Green lines (o-o) denote correct matches and red lines depict wrong matches (x-x).}
  \label{fig:sparse_matches_sup2}
\end{figure*}

We qualitatively compare our method with DINO+SD \cite{23_sd+dinov2} and GeoAware \cite{24_left_right} on all SPair-71k \cite{19_spair} categories in \cref{fig:sparse_matches_sup1} and \cref{fig:sparse_matches_sup2}. The results demonstrate that our method is applicable to a wide range of object-classes and can improve performance even if the 3D object-class representation is very coarse.

\section{Hyper-Parameters}
\label{sec:hyper_params}

We list the hyper-parameters in \cref{tab:parameters}. The hyper-parameters $\sigma_\text{dense}$, $\sigma_\text{sparse}$, and $\sigma_\text{sparse}^\text{inference}$ correspond to the $\sigma$ in \cref{e-spatial} for $C_\text{dense}$ and $C_\text{sparse}$, respectively. The hyper-parameter $\sigma_\text{sparse}^\text{inference}$ is used for sparse semantic correspondence in \cref{equ:rec_infer}. As discussed in the paper, it is crucial to start with a high $\sigma$ and decrease its value over time to obtain a good alignment. Furthermore, we generally decrease $\sigma_\text{dense}$ faster than $\sigma_\text{sparse}$ to avoid local minima. Additionally, we choose $\sigma_\text{sparse}^\text{inference}$ large for categories where our spatial prior is rather imprecise. However, as seen in \cref{t-spair-cat}, the imprecise spatial prior can still improve performance.

The weights $w_\text{dense}$ and $w_\text{sparse}$ are from \cref{e-reconstruct} and $w_\text{geom}$, $w_\text{background}$, and $w_\text{depth}$ are from \cref{equ:loss_align}. We start with $w_\text{sparse} = 0$ and increase its value over time to avoid local minima. Similarly, we also start with a small value for $w_\text{background}$ as this term can lead to divergence if the representation does not sufficiently overlap with the object instance in the image, which is the case at the beginning of the optimization.

\begin{table}[t]
    \centering
    \begin{adjustbox}{width=\linewidth}
    \begin{tabular}{llp{5cm}}
        \toprule
        \textbf{Group} & \textbf{Parameter} & \textbf{Value} \\
        \midrule
        \multirow{3}{*}{Optimizer}  & Type & AdamW \\
                                    & lr & 5e-3 \\
                                    & $n_\text{steps}$ & 1000 \\
        \midrule
        \multirow{7}{*}{Sigmas} & $\sigma_\text{dense}$ & Timesteps: [0, 300, 500] \\ 
                                &       & Values: [1.0, 0.1, 0.03] \\
                               & $\sigma_\text{sparse}$ & Timesteps: [0, 300, 500, 700] \\
                               &        & Values (Bottle): [1.0, 0.3, 0.3, 0.05] \\
                               &        & Values (Other): [1.0, 0.1, 0.1, 0.05] \\
                               & $\sigma_\text{sparse}^\text{inference}$ & Chair, TV: 0.01; Airplane, Bicycle, Bottle: 0.03; Other: 1.0 \\
        \midrule
        \multirow{12}{*}{Weights} & $w_\text{dense}$ & 1.0 \\
                                 & $w_\text{sparse}$ & Timesteps: [0, 500, 700, 1000] \\
                                 &        & Values (Bottle): [0, 0, 1, 10] \\
                                 &        & Values (Other): [0, 0, 10, 1] \\
                                 & $w_\text{geom}$ & 0.5 \\
                                 & $w_\text{background}$ & Bottle: \\
                                 &           & Timesteps: [0, 700, 900, 1000] \\
                                 &           & Values: [0, 0, 1, 20] \\
                                 &           & Other: \\
                                 &           & Timesteps: [0, 300, 500, 700] \\
                                 &           & Values: [1, 10, 100, 10] \\
                                 & $w_\text{depth}$ & 10.0 \\
        \bottomrule
    \end{tabular}
    \end{adjustbox}
    \caption{Hyper-Parameters. \textit{Values} are linearly interpolated according to \textit{Timesteps}.}
    \label{tab:parameters}
\end{table}

\end{document}